\documentclass[times,twocolumn,final]{elsarticle}

\usepackage{arxiv}
\usepackage{graphicx}%
\usepackage{multirow}%
\usepackage{amsmath,amssymb,amsfonts}%

\usepackage{amsthm}%
\usepackage{mathrsfs}%
\usepackage[title]{appendix}%
\usepackage{xcolor}%
\usepackage[utf8]{inputenc}
\usepackage{booktabs}%
\usepackage{algpseudocode,algorithm}
\MakeRobust{\Call}
\usepackage{pifont}
\usepackage{adjustbox}
\usepackage{graphicx}
\usepackage{float}
\usepackage{placeins}
\usepackage{stfloats}
\usepackage{fancyhdr}
\usepackage[hyphens]{url}
\usepackage{hyperref}
\usepackage{caption}
\usepackage{longtable}
\usepackage{pdfpages}

\usepackage{subcaption}

\usepackage{booktabs,tabularx,ragged2e,array,multirow,graphicx}
\newcolumntype{Y}{>{\RaggedRight\arraybackslash}X}

\fancypagestyle{firstpagestyle}{
    \fancyhf{} 
    \fancyhead{} 
    \fancyfoot{} 
}

\fancypagestyle{default}{
    \fancyhf{}
    \fancyhead[R]{\thepage} 
}

\begin{document}

\title{A Surgical Foundation Model Reveals Task-Dependent Label Efficiency}

\author[1,2,3]{Florian Philipp \snm{Stilz}\corref{corresp}}
\cortext[corresp]{Corresponding author: \texttt{florian.stilz@tum.de}}
\author[1,2]{Lorenzo \snm{Arboit}}
\author[1,2,4]{Vinkle \snm{Srivastav}}
\author[]{CAMMA International Surgical Partners}
\author[5]{Jacques \snm{Marescaux}}
\author[6]{Sergio \snm{Alfieri}}
\author[2,6]{Pietro \snm{Mascagni}}
\author[3]{Nassir \snm{Navab}}
\author[1,2]{Nicolas \snm{Padoy}}

\address[1]{University of Strasbourg, CNRS, INSERM, ICube, UMR7357, Strasbourg, France}
\address[2]{IHU Strasbourg, Strasbourg, France}
\address[3]{Technical University of Munich, Computer Aided Medical Procedures, Munich, Germany}
\address[4]{Indian Institute of Technology (IIT) Madras, Department of Data Science and AI, Wadhwani School of Data Science and AI (WSAI), Chennai, India}
\address[5]{IRCAD, Research Institute Against Digestive Cancer, Strasbourg, France}
\address[6]{Fondazione Policlinico Universitario Agostino Gemelli IRCCS, Rome, Italy}

\received{XXX}
\finalform{XXX}
\accepted{XXX}
\availableonline{XXX}
\communicated{XXX}

\begin{abstract}
Developing label-efficient models is a central challenge in surgical AI due to the high cost and scarcity of expert annotation. While self-supervised foundation models adapt well to new tasks with minimal data, how label efficiency varies across different surgical tasks remains largely unexplored. 

Here, we introduce SURGE, a surgical foundation model trained on SurgSpectrum-30M+, the largest pretraining dataset comprising over 30 million frames, with checkpoints released to enable further research. We systematically evaluate label efficiency across 5 task categories and 15 benchmarks. These range from temporal and spatial scene understanding to fine-grained reasoning tied to instrument–anatomy interactions and safety-critical maneuvers.

SURGE outperforms prior state-of-the-art on all benchmarks, even surpassing task-specific models on complex reasoning tasks. Crucially, we reveal a task-dependent scaling behavior: while scene understanding tasks saturate with minimal supervision, fine-grained reasoning tasks continue improving with substantially larger annotation budgets, providing a blueprint for allocating expert effort in
complex domains.

\textbf{Code}: \url{https://github.com/CAMMA-public/SURGE}.
\end{abstract}

\maketitle

\begin{figure*}
    \centering
    \vspace{-0mm}
    \includegraphics[width=1.00\linewidth]{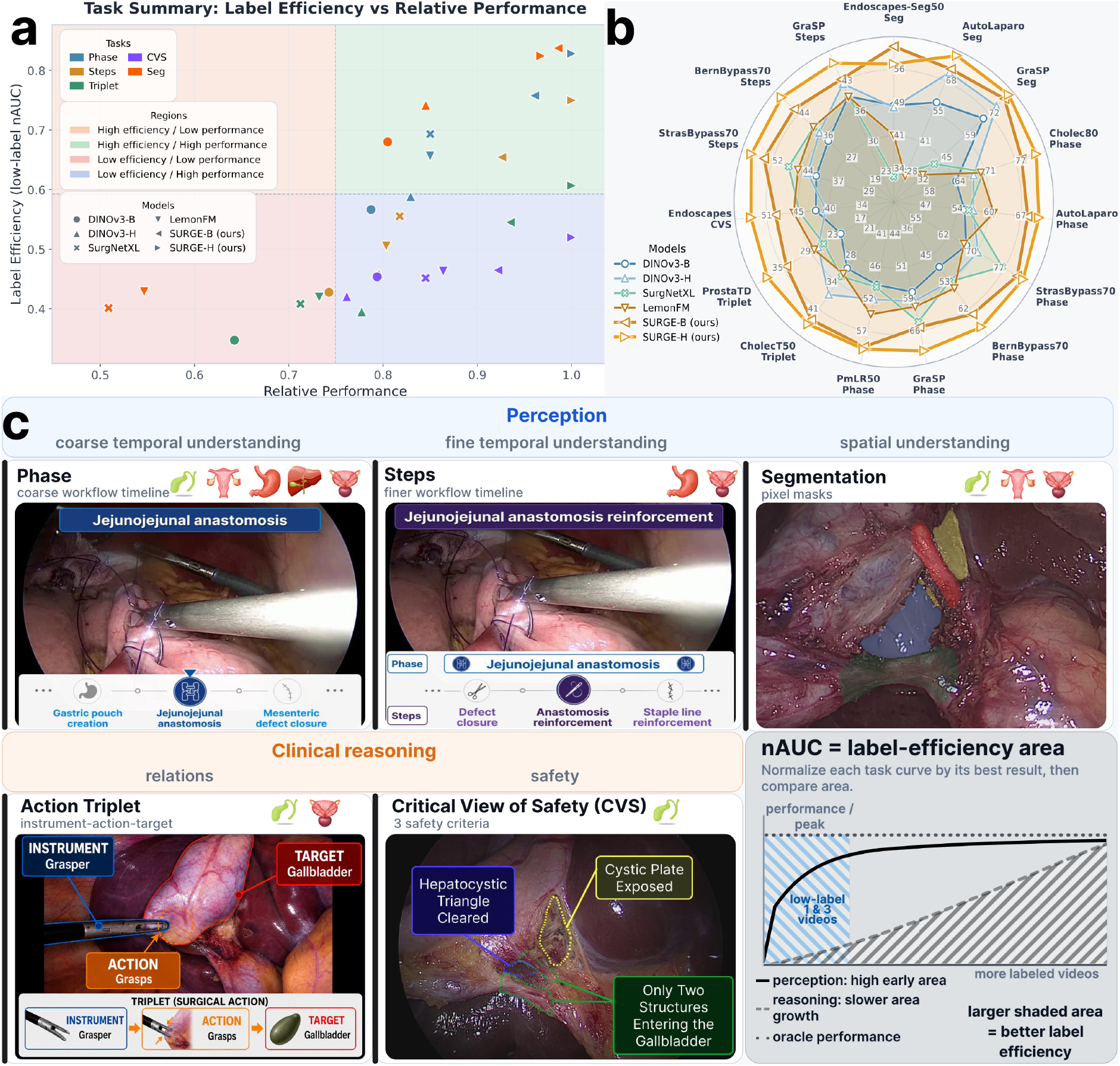}
    \captionof{figure}{
    \textbf{a}: Label efficiency across surgical tasks, shown as normalized area under the curve (nAUC) in low label regimes versus relative peak performance. Results are averaged across datasets per task. We observe a consistent ordering of tasks by data efficiency: perception tasks (segmentation (Seg), Phase, Steps) achieve strong performance in low-label regimes, while fine-grained and reasoning tasks (Action Triplet, CVS) require substantially more supervision. Notably, certain tasks exhibit low-data regimes with limited learning signal, indicating threshold effects that are not overcome by pretraining.
    \textbf{b}: Full-data performance comparison of our models with general-domain and surgical foundation baselines across 15 benchmarks.
    \textbf{c}: Five tasks are illustrated with representative examples and grouped into perception and clinical reasoning categories. Icons indicating anatomical structures denote the surgical procedures associated with each task (cholecystectomy, hysterectomy, liver resection, gastric bypass, prostatectomy). The nAUC metric is illustrated schematically. 
    }
    \label{fig:teaser}
\end{figure*}

\thispagestyle{firstpagestyle}

\section{Introduction}
\label{sec:introduction}

Artificial intelligence is increasingly deployed to support expert decision-making in complex, high-stakes domains, ranging from healthcare and science to engineering and public safety~\cite{sambasivan2021everyone}. In these settings, models are typically trained on annotated datasets that require substantial input from domain experts. However, generating such annotations is both time-consuming and costly, as expertise is scarce and tasks often demand deep contextual understanding. As a result, the development of new AI systems is frequently constrained not only by model capacity, but increasingly by the availability of high-quality labeled data~\cite{sambasivan2021everyone}. Improving the efficiency with which new tasks can be learned is therefore critical.
This can be approached from two complementary directions: developing models that achieve strong performance with fewer annotations and understanding how annotation requirements vary across tasks to allocate expert effort more effectively. Here, we address both challenges by developing a self-supervised foundation model for surgery and systematically characterizing how its performance scales with the amount of task-specific labeled data. We use this scaling behavior to quantify label efficiency across surgical perception and reasoning tasks, providing a principled basis for understanding and reducing the annotation requirements of AI systems in surgery.

Surgical procedures are complex, high-risk interventions that require continuous perception and timely reasoning to support safe and effective decision-making. Automated analysis of endoscopic video acquired during minimally invasive surgery has the potential to improve intraoperative guidance, surgical training and postoperative assessment. Surgical AI~\cite{maier2022surgical,maier2017surgical,vedula2017surgical,mascagni2022computer} is therefore a particularly high-stakes application, as errors and preventable complications during surgery contribute substantially to global morbidity and mortality~\cite{nepogodiev2019global,zegers2011incidence}, highlighting the need for reliable intraoperative decision support.
Importantly, surgical video analysis encompasses multiple levels of understanding, ranging from temporal perceptual tasks, such as recognizing surgical phases and steps~\cite{twinanda2016endonet,wang2022autolaparo,lavanchy2024challenges,ayobi2025pixel,ye2026self} and spatial tasks such as segmenting instruments and anatomy~\cite{murali2023endoscapes,allan20192017,wang2022autolaparo,ayobi2025pixel,kamtam2025fine,kamtam2025surgisam2,de2026surgical}, to higher-level reasoning about instrument–anatomy interactions~\cite{nwoye2022rendezvous,chen2025prostatd} and clinically grounded decision-making~\cite{murali2023endoscapes}. These diverse tasks vary substantially in complexity, raising the question of whether they also differ in the amount of supervision required to learn them effectively.

Recent advances in self-supervised and foundation model pretraining~\cite{chen2020simclr,zhou2022mugs,zhou2021ibot,simeoni2025dinov3} have enabled strong general-purpose visual representations by learning from large amounts of unlabeled data. When applied to surgical video, these approaches often yield competitive performance, particularly when sufficient labeled data is available~\cite{ramesh2023dissecting,batic2024endovit,de2025scaling}. Domain-adapted pretraining~\cite{ramesh2023dissecting,batic2024endovit,jong2025gastronet,de2025scaling,che2026lemon,de2026towards} and multimodal approaches~\cite{yuan2025learning,stilz2026clipper} have further improved data efficiency. However, existing work has primarily focused on developing stronger representations, while the extent to which different categories of surgical tasks exhibit distinct label efficiency remains largely unexplored.

In this work, we address this gap through a large-scale, systematic analysis of label efficiency in surgical AI. To this end, we introduce SURGE (\textbf{S}urgical \textbf{U}nderstanding through \textbf{R}epresentation and \textbf{G}eneralization at Scal\textbf{e}), a surgical foundation model trained via self-supervised learning on SurgSpectrum-30M+, a dataset of over 30 million surgical video frames. Using this unified representation, we evaluate performance across a spectrum of tasks, which we group into perceptual tasks (e.g., spatial and temporal understanding) and clinically grounded reasoning tasks (e.g., relational instrument–anatomy interactions and safety-critical maneuver recognition), which require higher-level judgment built upon perceptual understanding. Beyond overall performance, we study how these tasks scale with increasing supervision. While large-scale domain-adapted pretraining substantially reduces labeled training data requirements, it does not homogenize performance. Instead, tasks exhibit distinct label efficiencies that persist even under stronger supervision and higher-parameter adaptation, suggesting that the efficiency with which different aspects of surgical understanding can be learned depends not only on representation quality, but also on the structure of the task itself.
\section{Results}\label{results}

Understanding label efficiency in surgical computer vision requires disentangling the interplay between representation quality, task structure, and supervision.

To this end we build on large-scale domain-adapted self-supervised pretraining, training $\mathrm{ViT}{\text{-Base}}$ and $\mathrm{ViT}{\text{-Huge+}}$ on SurgSpectrum-30M+ (over 30 million surgical video frames), resulting in SURGE-B and SURGE-H, providing a strong and unified visual representation for surgical video analysis. As a result, SURGE provides a well-suited foundation for studying label efficiency, as it reduces performance bottlenecks associated with weaker or less robust representations. This unified representation enables us to study supervision scaling while reducing confounding effects from representation quality.

We evaluate these representations across a diverse set of surgical tasks, spanning two main categories: surgical perception and more fine-grained and clinically grounded reasoning tasks. Surgical perception tasks are separated into temporal workflow understanding at different granularities (phase and step recognition) and spatial perception (semantic segmentation), while reasoning tasks focus on analyzing surgical execution. These reasoning tasks include clinically grounded assessments (Critical View of Safety recognition)~\cite{strasberg2010rationale} as well as relational instrument-anatomy understanding (action triplet recognition), which has shown correlations with patient outcomes~\cite{heard2026ai}. Action triplet recognition refers to predicting surgical actions as $\langle \text{instrument}, \text{verb}, \text{target} \rangle$ tuples 
of tool-tissue interactions.

We systematically vary the amount of supervision, from a single training video to the full training set for each task, and evaluate all models using a unified linear-probing protocol in which only a linear classifier is trained on frozen representations. This protocol isolates the contribution of representation quality to downstream label efficiency. We compare SURGE with general-domain foundation models~\cite{simeoni2025dinov3} and prior surgical self-supervised approaches~\cite{ramesh2023dissecting,che2026lemon,de2025scaling}. Across tasks, performance improves consistently with increasing supervision, while stronger representations provide substantial gains, particularly in low-label regimes.

We quantify label efficiency using normalized area under the curve (nAUC; see Methods and see Fig.~\ref{fig:teaser}c). We compute nAUC in two settings: over the full training budget and in a low-label regime consisting of 1 and 3 training videos.

\subsection{SURGE outperforms other foundation models across surgical AI tasks}

\begin{figure}
    \centering
    \includegraphics[width=1.0\linewidth]{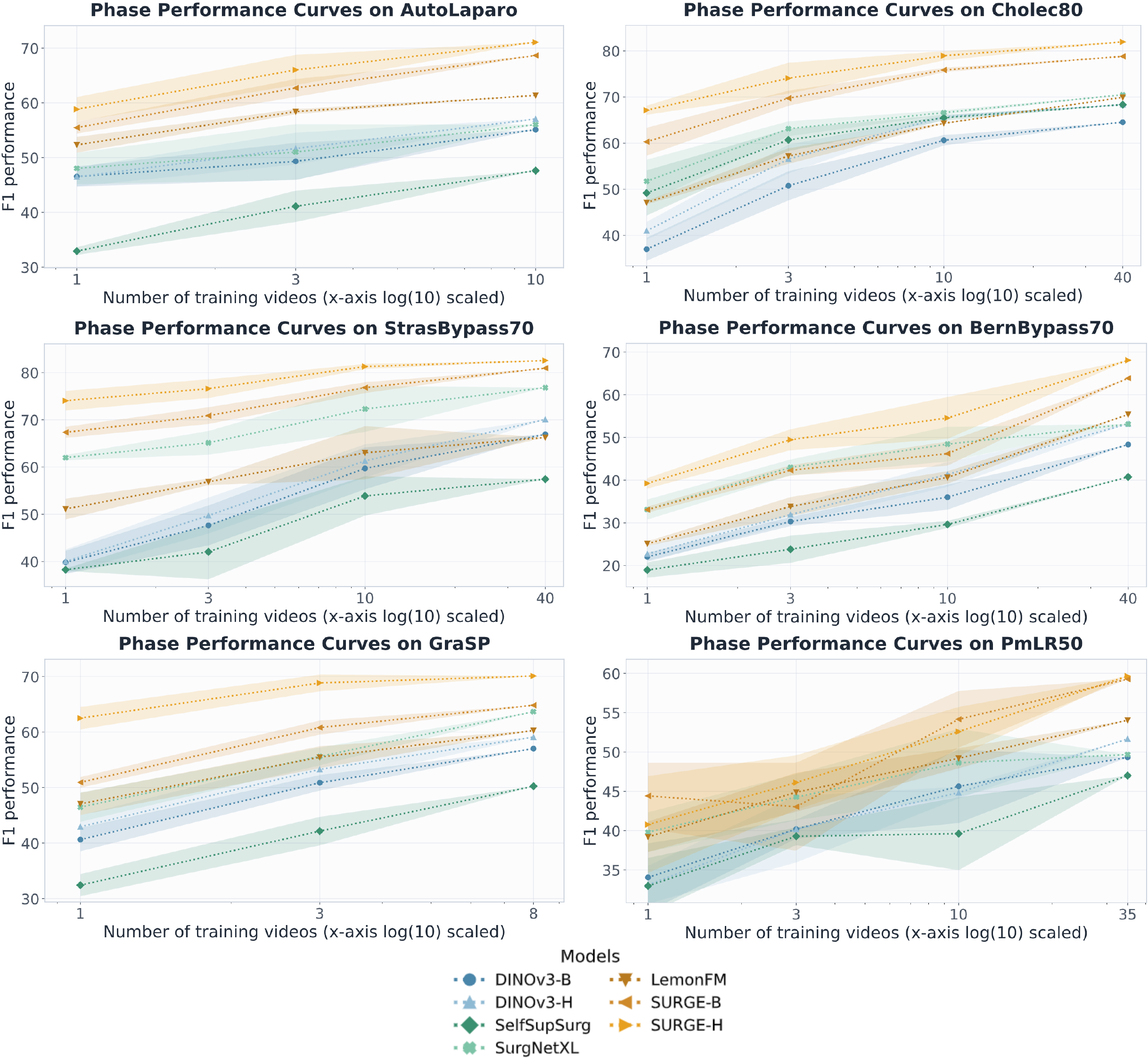}
    \caption{\textbf{Coarse temporal perception (phase recognition) performance curves}: All subset runs are averaged over 3 runs with error bands highlighting the standard deviation.}
    \label{fig:phase}
\end{figure}

\begin{figure}
    \centering
    \includegraphics[width=1.0\linewidth]{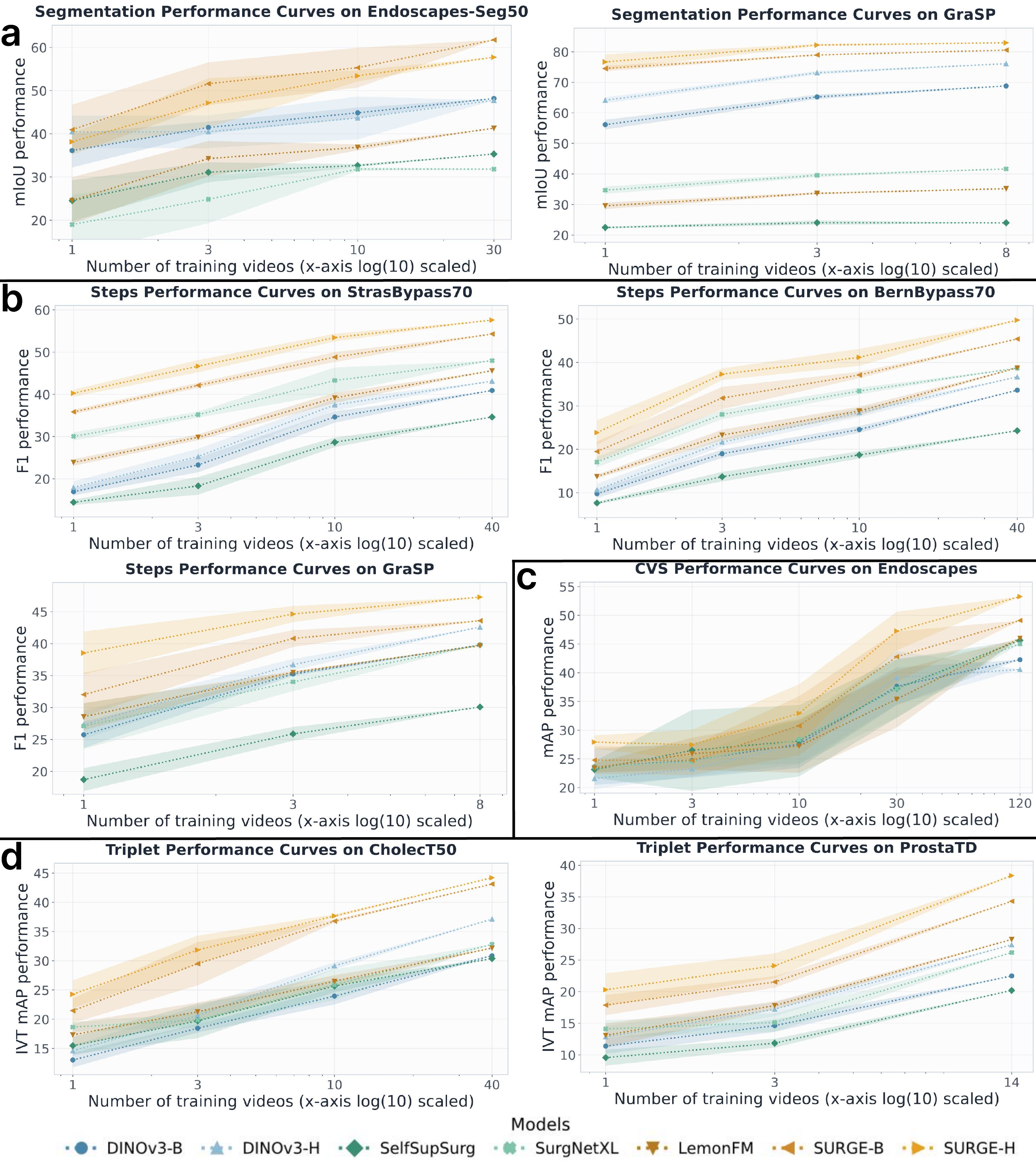}
    \caption{\textbf{Additional perception \& reasoning task performance curves}: \textbf{a} shows spatial understanding (semantic segmentation), \textbf{b}: fine temporal understanding (step recognition), \textbf{c}: safety maneuver analysis (CVS), and in \textbf{d} instrument-anatomy interaction (action triplet). All subset runs are averaged over 3 runs with error bands highlighting the standard deviation.}
    \label{fig:curve_big}
\end{figure}

SURGE achieves the best performance across all 15 benchmarks spanning diverse surgical tasks and supervision regimes (Fig.~\ref{fig:teaser}, Figs.~\ref{fig:phase},~\ref{fig:curve_big}). For the subsequent label-efficiency analysis, we consider 14 benchmarks for which supervision can be systematically varied at the video level. AutoLaparo-Seg~\cite{wang2022autolaparo} is evaluated using its official train/validation/test splits only, as its data cannot be partitioned into individual videos for controlled supervision scaling. We illustrate this overall advantage with representative results from temporal workflow understanding and fine-grained relational reasoning.

For temporal workflow understanding, SURGE-B surpasses the strongest baseline on every dataset, achieving an average F1 score of 69.40 (95\% CI: 63.30-75.85), compared with 63.63 (95\% CI: 57.33-70.08) for the strongest competing foundation model. Scaling the representation from SURGE-B to SURGE-H further improves performance to 72.23 (95\% CI: 66.18-78.51), demonstrating consistent gains with increased model capacity.

The advantage extends beyond temporal perception to more fine-grained relational reasoning. On instrument-anatomy interaction recognition, SURGE-B achieves 38.72 mAP (95\% CI: 34.3-43.13), compared with 32.87 (95\% CI: 28.6-37.13) for the strongest per-dataset baselines. SURGE-H further increases performance to 41.40 mAP (95\% CI: 38.6-44.2). Thus, domain-adapted pretraining consistently improves surgical video representations across both temporal perception and fine-grained relational understanding, with further gains from scaling model capacity.

\subsection{Consistently superior label efficiency across all annotation regimes}

\begin{figure*}
    \vspace{-3mm}
    \centering
    \vspace{-17mm}
    \includegraphics[width=0.8\linewidth]{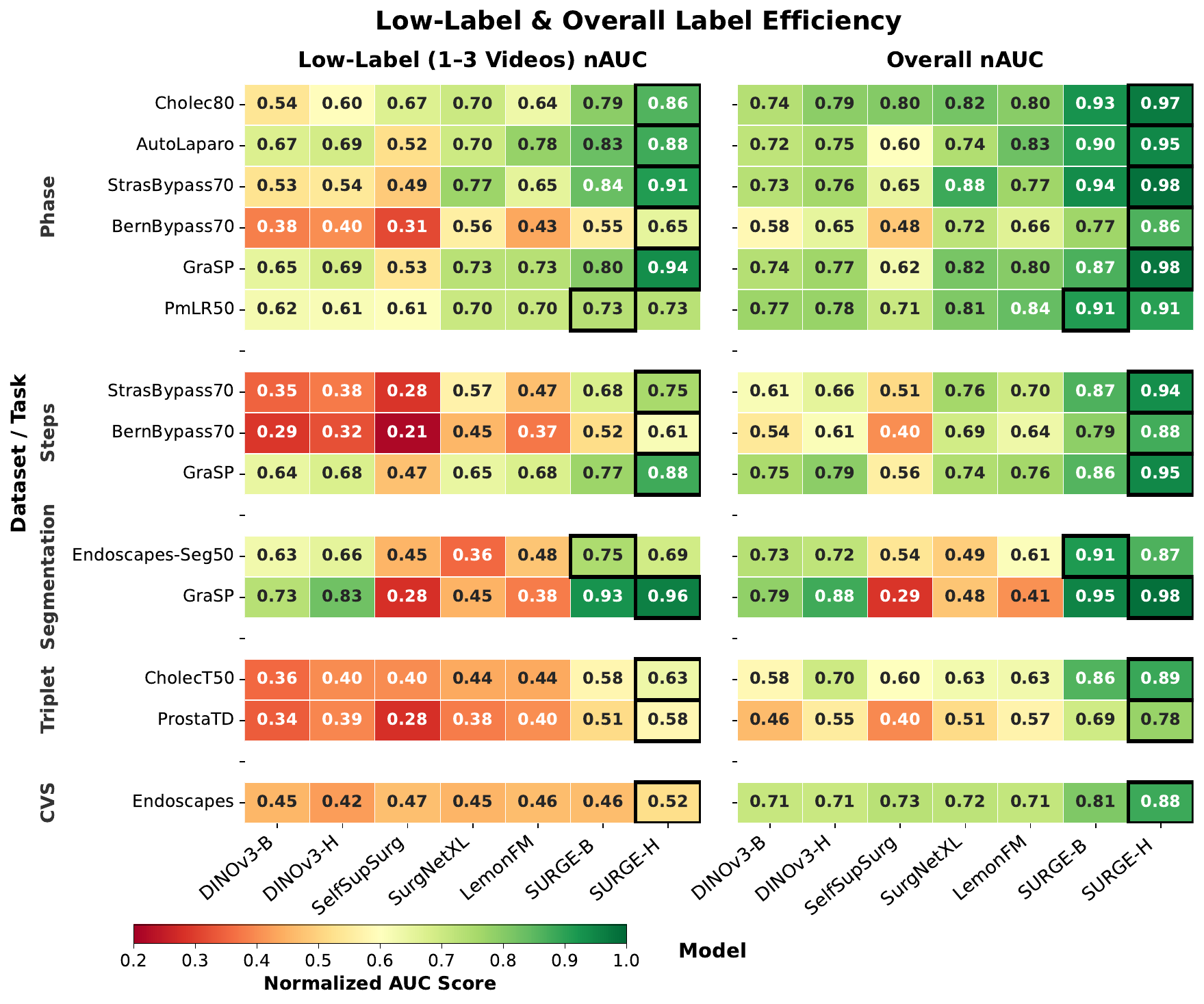}
    \caption{\textbf{Overall and low-data regime label efficiency comparison}: measuring normalized AUC (nAUC) across tasks/datasets, highlighting most efficient model both in heatmap color coding and box highlighting.}
    \label{fig:auc}
\end{figure*}

We next examine whether the performance gains of SURGE translate into greater label efficiency across the full range of annotation budgets (Fig.~\ref{fig:auc}).
Averaged across datasets, SURGE-H achieves an nAUC of 0.914 (95\% CI: 0.884–0.942), substantially outperforming the strongest baselines on each dataset with 0.769, (95\% CI: 0.724–0.809). This advantage persists in low-label settings (Fig.~\ref{fig:auc}), where SURGE-H attains a low-label nAUC of 0.757 (95\% CI: 0.683–0.835) versus 0.624 (95\% CI: 0.549–0.692).

We further analyze the effect of model scale. SURGE-B and SURGE-H contain 85.67M and 840.63M parameters, respectively, and larger models consistently improve performance, with SURGE-H outperforming SURGE-B in 14 out of 15 benchmarks, as depicted in Fig.~\ref{fig:teaser}b. This improvement is especially pronounced in low-label settings; while task variance results in wide confidence intervals, paired analysis reveals a highly significant and substantial performance gain for the larger model (SURGE-H: 0.757 (95\% CI: 0.683–0.835) and SURGE-B: 0.696 (95\% CI: 0.619-0.769)); Wilcoxon signed-rank $p < 0.01$, Cohen's $d = 1.25$).

Overall, SURGE substantially improves label efficiency in surgical AI, enabling strong performance with reduced annotation effort. These gains are consistent across five surgical procedures (Fig.~\ref{fig:auc}), although the magnitude of improvement varies across tasks, reflecting differences in how performance scales with supervision.

\subsection{Domain adaptation bridges general and surgical representation strengths}

We next examine why SURGE performs consistently across diverse surgical tasks. Prior surgical self-supervised approaches are particularly effective on workflow and clinically grounded tasks (Critical View of Safety), especially in low-data regimes, but are less effective for dense spatial perception and fine-grained instrument-anatomy interaction semantics. In contrast, general-domain self-supervised models provide stronger representations for tasks such as semantic segmentation and instrument-anatomy interaction (Fig.~\ref{fig:auc}). 

SURGE combines these complementary strengths. It retains the strong performance of surgical self-supervised approaches on workflow and clinically grounded tasks while exceeding general-domain models on dense spatial and interaction-level tasks. For example, SURGE-H achieves 82.95 mIoU on GraSP, compared with 76.09 for DINOv3-H, and 44.20 mAP on CholecT50, compared with 37.13 for DINOv3-H. These results suggest that large-scale domain adaptation enables representations that capture both general visual structure and fine-grained surgical semantics, providing a potential basis for SURGE's broad transferability and label efficiency.

\subsection{Perception tasks exhibit high label efficiency with task structure dependent variations}

Our analysis reveals that surgical perception tasks exhibit high label efficiency, with performance saturating under minimal supervision. Tasks that rely primarily on visual cues or coarse temporal structure achieve strong performance with very limited annotated data, with diminishing gains as more data is added. For instance, in semantic segmentation on prostatectomy (GraSP dataset), SURGE-H attains an mIoU of 76.68 $\pm$ 2.53\% using only a single training video, with performance remaining nearly unchanged as the training set scales to eight videos as shown in Fig.~\ref{fig:curve_big}a. 
Similarly, in phase recognition, SURGE-H reaches 74.04 $\pm$ 2.09\% F1 on gastric bypass (StrasBypass70) with one video and quickly plateaus with additional data, while achieving comparable behavior on cholecystectomy (Cholec80) depicted in Fig.~\ref{fig:phase}. In these settings, a single labeled video often suffices to outperform competing pretrained surgical models trained on substantially larger datasets.

While this regime of high label efficiency is consistent across perception tasks, we find that structural properties of the task modulate the rate at which saturation is reached. Increasing the granularity of the prediction objective introduces a more pronounced dependence on labeled data. On StrasBypass70, while coarse phase recognition improves modestly with additional data (8.5\% F1 gain from 1 to 40 videos), the more fine-grained step recognition task exhibits a substantially steeper scaling curve, improving by 17.3\% F1 over the same range seen in Fig.~\ref{fig:curve_big}b.
Thus, increasing temporal granularity substantially increases the amount of supervision required to reach comparable performance.

Similarly, the complexity of the annotations influences data requirements even within this otherwise efficient regime. Comparing semantic segmentation across two datasets (GraSP and Endoscapes-Seg50) in Fig.~\ref{fig:curve_big}a 
illustrates this effect. GraSP reaches strong performance quickly with limited data, whereas Endoscapes-Seg50 improves more gradually, starting from a lower baseline and continuing to benefit from additional annotations. This difference reflects the nature of the labels: GraSP focuses on a smaller set of well-defined structures, primarily surgical instruments, whereas  
Endoscapes-Seg50 includes a larger and more fine-grained label space, including small anatomical structures alongside instruments, resulting in a more complex prediction problem.
As a result, the model must learn a more complex mapping from visual features to labels, which increases the amount of data required to achieve comparable performance.

Taken together, these results show that surgical perception tasks form a regime of high label efficiency characterized by rapid saturation with minimal supervision. Although factors such as task granularity, annotation density, and semantic complexity modulate the precise label efficiency, they do not alter the overall trend of efficient learning in these tasks. 
In the next section, we show that this regime breaks down for reasoning tasks, 
which exhibit distinctly different label efficiency and remain far from saturation even under substantially increased supervision.

\subsection{Reasoning tasks exhibit persistent data requirements}

In contrast to the rapid saturation observed in perception tasks, reasoning tasks exhibit fundamentally different label efficiency, remaining far from saturation even under substantially increased supervision.

In the instrument-anatomy interaction tasks (action triplet recognition) on the CholecT50 and ProstaTD benchmarks shown in Fig.~\ref{fig:curve_big}d
, SURGE-H maintains a consistent lead over all baselines, yet displays a remarkably persistent log-linear scaled curve that still allows for sustained performance gains under constant increase in labeled data. For instance, on ProstaTD, SURGE-H begins at 20.30 $\pm$ 2.53 mAP and requires 14 training videos to reach 38.36 mAP, a trajectory that lacks the 'jump-start' characteristic, where 1-video performance is already a significant fraction of the ceiling, observed in temporal (phase) or spatial (semantic segmentation) perception tasks. This is also supported by the nAUC metric, where perception tasks achieve average scores of 0.925, 0.940, and 0.919 for semantic segmentation, phase recognition, and step recognition, respectively, compared to 0.832 for action triplet recognition. This difference is also present in the low-label regime, where semantic segmentation, phase recognition, and step recognition achieve average low-label nAUC values of 0.825, 0.828, and 0.750, respectively, versus 0.607 for instrument-anatomy interaction (action triplet recognition).

This structural bottleneck is most pronounced in the clinical quality assessment (Critical View of Safety (CVS)), depicted in Fig.~\ref{fig:curve_big}c, 
on the Endoscapes dataset. Despite leveraging the same large-scale pretrained representation, performance for SURGE-H starts at 27.92 $\pm$ 1.17 mAP and requires substantially more labeled data, reaching 53.26 mAP only when scaling to 120 training videos. In contrast to other tasks, we observe an apparent threshold-like behavior, where all models exhibit limited learning signal when trained with 10 or fewer videos, and only begin to improve meaningfully beyond this regime. This pattern is further reflected in the label-efficiency metric (nAUC), which reaches only 0.882 when evaluated across all training budgets and drops to 0.520 in the low-label regime. Notably, the low-label nAUC is substantially lower than that observed for any other task, indicating that this task benefits less from limited supervision. This suggests that certain reasoning tasks require a minimum amount of supervision before meaningful generalization can occur.

\subsection{Label instances as an indicator of supervision requirements}

\begin{figure*}
    \centering
    \vspace{-17mm}
    \includegraphics[width=0.8\linewidth]{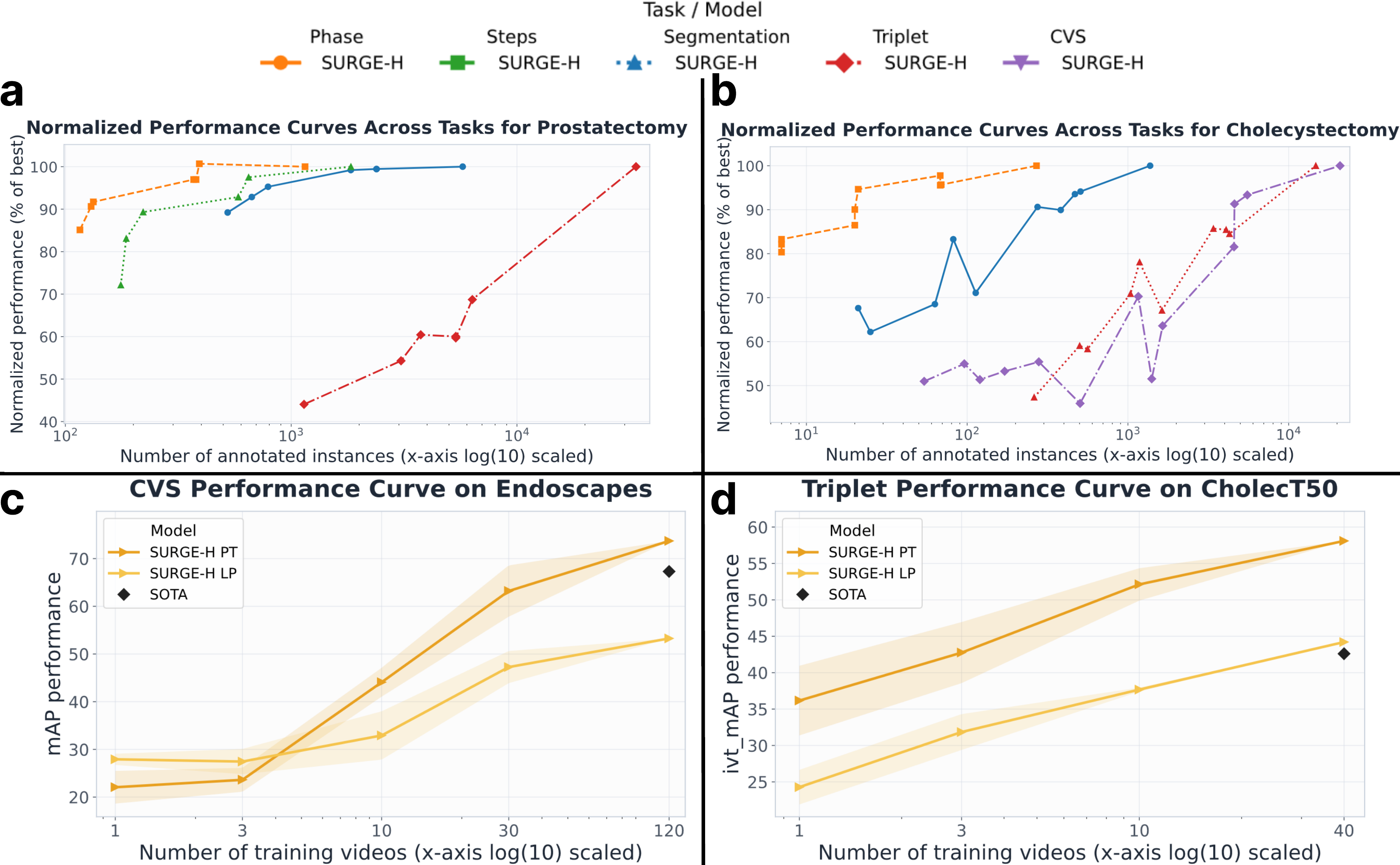}
    \caption{\textbf{Label instances and label efficiency comparison across tasks \& higher parameter adaptation results}: \textbf{a} shows task comparisons for prostatectomy and \textbf{b} for cholecystectomy. Highlighting severe annotation density differences across tasks. \textbf{c}\&\textbf{d}: We select the best model for this analysis (SURGE-H) and compare linear probing (SURGE-H LP) and partial tuning of the backbone (SURGE-H PT). All subset runs are averaged over 3 runs with error bands highlighting the standard deviation. SOTA on Endoscapes is LG-CVS~\cite{murali2023latent} and on CholecT50 is TripletGCN~\cite{zou2025capturing}.}
    \label{fig:label_occ_sota}
\end{figure*}

To further ground these differences in data requirements, we next quantify the underlying supervision effort associated with each task, providing a practical proxy for label efficiency. While true annotation cost is a complex composite of mechanical time (e.g., pixel-wise segmentation) and cognitive load (e.g., clinical consensus for safety criteria), direct measurement of human hours across retrospective, multi-center datasets is very difficult and annotator specific. 
To overcome this limitation and provide a standardized baseline, we introduce label instances, the absolute count of all explicitly annotated instances across a dataset, as an objective, macro-level proxy for total annotation effort. By systematically quantifying the volume of human-generated labels required per task category, we offer a first-of-its-kind approximation of comparative supervision demands. As shown in Fig.~\ref{fig:label_occ_sota}a-b, this resulting distribution closely mirrors the previously observed scaling patterns, demonstrating that total label instances serves as a powerful indicator of the intrinsic human effort required to solve different surgical tasks.

Notably, more clinically relevant tasks requiring clinical reasoning and relational instrument-anatomy understanding (e.g., CVS and action triplet recognition) exhibit substantially higher label counts, reflecting significantly greater annotation effort. Despite this increased supervision, their performance continues to improve with additional annotations without clear signs of saturation. This indicates that these tasks remain data-hungry even under substantially higher annotation volumes.

In contrast, temporal and spatial perception tasks rely on comparatively fewer annotations and exhibit more favorable learning efficiency. Among these, semantic segmentation stands out with the highest number of labels per video due to its dense annotation structure, yet still benefits from comparatively efficient learning dynamics.

Overall, these findings reveal a pronounced imbalance in supervision demands across tasks and indicate that label efficiency is strongly task-dependent.

\subsection{Higher parameter adaptation surpasses supervised SOTA without altering label efficiency}
\label{higher-param}

Given the strong task-dependent differences in label efficiency observed above, we next examine whether these dynamics can be altered through higher parameter adaptation, in which more parameters of the pretrained backbone are updated using task-specific supervision rather than training only a linear classifier. While higher parameter adaptation substantially improves absolute performance, enabling data-intensive tasks such as relational instrument-anatomy recognition and clinical reasoning to surpass task-specific supervised state-of-the-art baselines, the fundamental label efficiency with respect to supervision remains unchanged.

As shown in Fig.~\ref{fig:label_occ_sota}c-d, partial adaptation consistently shifts performance upwards, yet preserves the relative gains obtained from additional supervision. Notably, this upward shift yields absolute state-of-the-art gains of 6.4\% mAP on Endoscapes and 15.5\% mAP on CholecT50 action triplet recognition. These improvements are achieved using our generalized approach, without relying on the complex, task-specific modifications used by previous baselines, such as segmentation-based supervision~\cite{murali2023latent} or ensemble strategies~\cite{zou2025capturing}.
\section{Discussion}\label{discussion}

Our results provide a systematic view of how surgical AI learns from expert supervision across tasks with fundamentally different demands. Large-scale domain-adapted self-supervised pretraining produces a broadly transferable representation that consistently outperforms both general-domain and prior surgical foundation models across 15 benchmarks, while providing substantial gains in low-label regimes. However, our results also reveal that representation quality alone does not determine label efficiency. Surgical perception tasks typically reach high performance with minimal supervision and rapidly saturate, whereas clinically grounded and relational reasoning tasks continue to benefit from substantially more supervision. Increasing the number of adapted parameters raises absolute performance and enables state-of-the-art results on these more demanding tasks, but does not fundamentally alter their scaling with supervision. Together, these findings suggest that representation quality determines how effectively models can exploit available supervision, whereas the structure of the downstream task determines how much supervision is ultimately required.

\subsection{The saturation of perception tasks}

The rapid saturation observed in spatial understanding tasks (semantic segmentation) and temporal understanding (phase recognition) suggests that the field may be approaching a milestone in perceptual learning. Our results show that with sufficiently large domain-adapted pretraining, the model’s internal representation of surgical instruments, tissue textures, and broad procedural states becomes ``task-ready" with very limited label supervision. In these regimes, a single training video acts not as a source of feature learning, but as a ``Rosetta Stone" that simply maps existing high-quality representations to class names. 

This finding suggests an immediate pivot in how we allocate annotation resources. For standard procedures and instrument-centric tasks, the pursuit of larger datasets may exhibit diminishing returns. Instead, these ``solved" perceptual components should be leveraged as automated foundations to assist in the more difficult task of mapping complex medical/surgical logic.

\subsection{The persistent reasoning gap and the need for hybrid modeling}

In contrast to the saturation of perception, fine-grained and reasoning tasks such as action triplet recognition and Critical View of Safety (CVS) assessment exhibit a “Reasoning Gap” that is not fully addressed by scale alone. These tasks require the model to move beyond \textit{identification} toward \textit{compositional logic} and \textit{clinical judgment}. The persistent growth of the scaling curves observed in these domains indicates that the model is actively learning the underlying rules of surgery frame-by-frame, rather than solely relying on pretrained visual archetypes. 

This resistance to pretraining gains does not suggest a limitation of the underlying representations, adaptation strategy, or model capacity, but rather may reflect the current frontier of deep 
learning. To cross this Reasoning Gap, we appear to require substantially more annotated data efforts compared to prior perception tasks or a transition toward hybrid methodologies. Such approaches might include a combination of largely-pretrained foundation models with task-specific architectural modeling, the integration of multi-modal clinical inputs, or the development of pretraining objectives that explicitly target causal and relational structures to learn those complicated problems under limited supervision efficiently. 

\subsection{A strategic roadmap to jumpstart Surgical AI}

By characterizing how data requirements vary across surgical tasks, our work provides a practical framework for understanding their label efficiency and guiding annotation strategies. This perspective enables a more principled allocation of labeling effort toward clinically impactful objectives:

\begin{itemize}
    \item \textbf{Low-shot transfer for perception:} Tasks such as instrument segmentation and coarse video understanding (phase recognition) exhibit high label efficiency and benefit strongly from domain-adapted pretraining. For these tasks, annotation requirements can be substantially reduced, opening the door to few-shot or weakly supervised approaches.
    
    \item \textbf{Targeted annotation for reasoning:} In contrast, tasks involving relational instrument-anatomy understanding and clinical assessment remain significantly more data-intensive. Expert annotation effort, being scarce and costly, should therefore be prioritized toward these settings.
    
    \item \textbf{Toward hybrid surgical AI:} Bridging the gap between perception and higher-level understanding is likely to require approaches that go beyond representation learning, combining strong visual foundations with task-aware modeling, temporal reasoning, and integration of clinically relevant context. Such approaches may also be important for achieving the robustness required for deployment in real-world clinical settings.
\end{itemize}

Ultimately, our findings suggest that while foundation models have substantially advanced the visual perception capabilities of AI, translating these representations into data-efficient reasoning remains a key challenge. By grounding this perspective in task-dependent data requirements, we outline a path toward more efficient, scalable, and clinically meaningful AI systems.

\subsection{Limitations \& Future Work}

While our study provides a controlled analysis of representation quality through linear probing and partial fine-tuning, this protocol inherently underestimates the peak performance achievable through task-specific fine-tuning. To realize its full potential, we release this foundation model as a shared tool for the research community. Future optimization, incorporating this model into specialized pipelines via architectural adapters or proxy losses, will enable practitioners to deeply adapt these baseline representations for their own specialized clinical applications.

A more fundamental limitation lies in the ``Reasoning Gap" identified in our scaling analysis. The persistent log-linear data requirements for CVS and action triplet recognition suggest that distillation-based pretext tasks may not fully capture the causal and logical dependencies of surgery. To resolve this, future pretraining should move toward embedding a deeper clinical reasoning behind observed videos and multimodal foundations, specifically targeting vision-text alignment (e.g., surgical reports). Such models could enable zero-shot transfer and may facilitate semi-automated annotation in label-sparse environments, effectively ``jump-starting" the development of AI for rare or specialized medical procedures.
\section{Methods}\label{methods}

\subsection{SurgSpectrum-30M+: pretraining dataset generation}
\label{sec:pretrain_data}

\begin{figure*}
    \centering
    \includegraphics[width=0.95\linewidth]{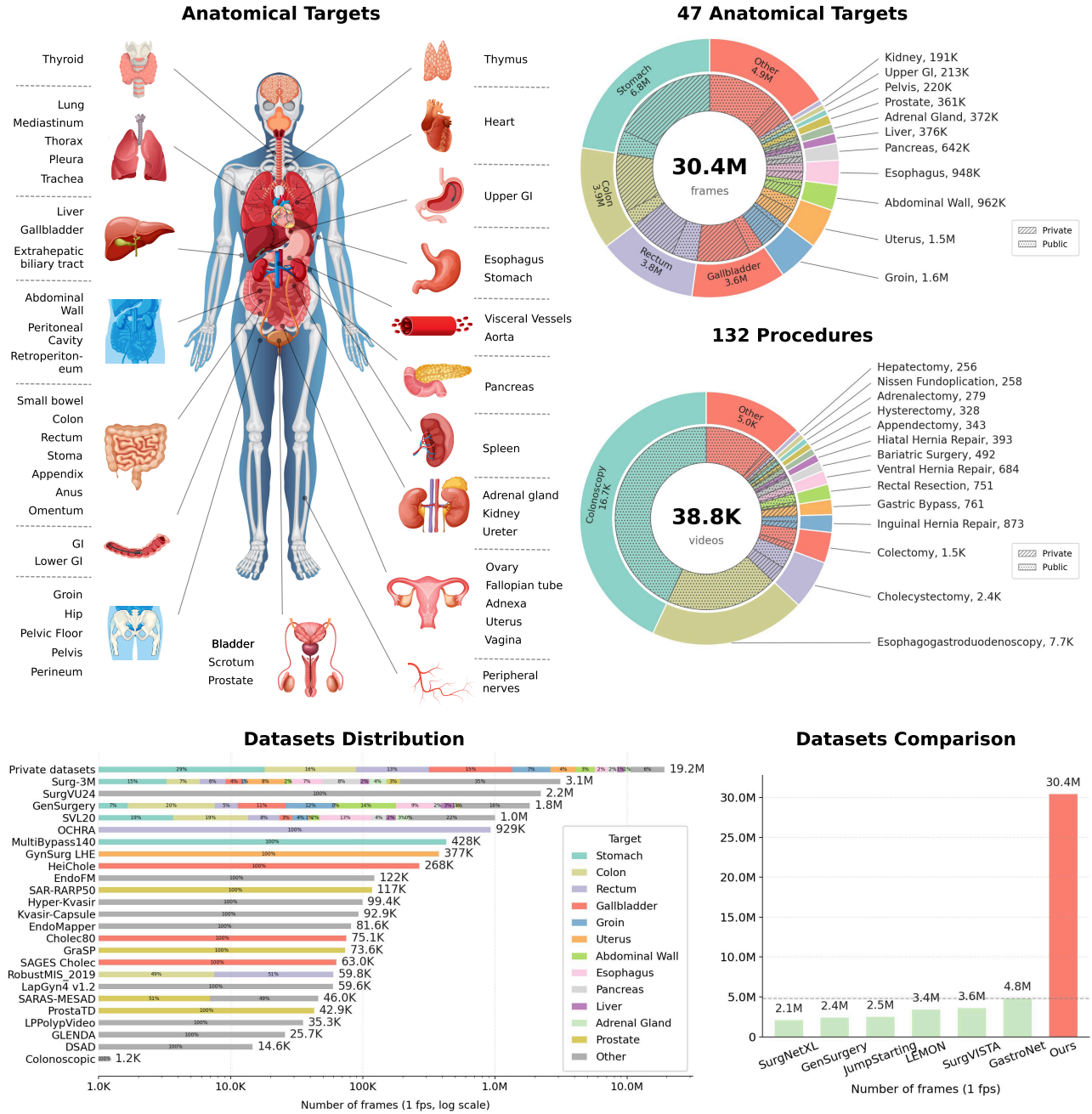}
    \caption{
    \textbf{SurgSpectrum-30M+.} Overview of the proposed large-scale surgical pretraining dataset. Dataset composition is visualized on an absolute scale for total size, while source distributions are reported as percentages. Compared to existing resources, SurgSpectrum-30M+ increases data scale by more than 6$\times$ relative to the next largest dataset, comprising over 30 million frames from approximately 39k videos. The dataset spans more than 130 surgical procedures, substantially increasing procedural diversity and enabling large-scale representation learning in the surgical domain.
    }
    \label{fig:data}
\end{figure*}

In order to unlock a truly label-efficient and multi-procedure foundation model, we construct \textbf{SurgSpectrum-30M+}, the largest surgical pretraining corpus to date. The dataset is assembled from a combination of public and proprietary sources across multiple centers and is designed to capture broad variability across surgical practice, spanning 132 procedures and 47 anatomical targets; more details can be seen in Fig. ~\ref{fig:data}. We curated this dataset from multiple public sources~\cite{twinanda2016endonet,schmidgall2024general,ayobi2025pixel,chen2025prostatd,azagra2023endomapper,carstens2023dresden,DBLP:conf/mmm/LeibetsederKSKK20,wagner2023comparative,borgli2020hyperkvasir,smedsrud2021kvasir,DBLP:conf/mmsys/LeibetsederPPKM18,lavanchy2024challenges,wang2023foundation,ghamrawi2025rectal,ross2020robust,che2025surg,psychogyios2023sar,zia2025surgical,ma2021ldpolypvideo,bawa2021saras,alapatt2025sages,nasirihaghighi2025gynsurg,yuan2025learning} as well as in-house surgical recordings and sampled frames from each video at 1 FPS which ensures a certain degree of visual diversity. To avoid any data leakage for potential downstream tasks we ensured to remove all videos/frames associated to validation or test sets of the public sources. As a critical preprocessing step, we train an out-of-body detector to remove non-informative content, including preparation phases and external camera views, thereby retaining only intraoperative endoscopic footage. This filtering step reduces visual noise and enforces domain specificity, resulting in a dataset that emphasizes clinically meaningful patterns while preserving diversity. The final corpus comprises \textbf{30,381,943 frames from 38,753 videos}.
\subsection{Model architecture}

The network employed in this work is based on the transformer architecture~\cite{vaswani2017attention}, which leverages self-attention to model long-range dependencies and capture global contextual interactions. Originally developed for natural language processing, transformer-based models have since demonstrated strong performance in visual representation learning by operating on images as sequences of patches~\cite{dosovitskiy2020image}. This formulation enables the encoding of complex spatial and semantic relationships within visual data.

Specifically, input images are partitioned into non-overlapping patches of size $16 \times 16$, which are flattened and projected into a learnable embedding space. Positional information is incorporated using rotary positional embeddings (RoPE)~\cite{su2021roformer}, allowing the model to capture relative spatial relationships between patches. The resulting sequence of patch embeddings is processed by a stack of transformer layers, yielding both global image-level representations and localized patch features.

We consider two model scales based on the Vision Transformer (ViT) design: $\mathrm{ViT}{\text{-Base}}$ and $\mathrm{ViT}{\text{-Huge+}}$. Both models are initialized from pretrained DINOv3 checkpoints and subsequently adapted to our surgical pretraining dataset. The base model comprises 12 transformer layers with a hidden dimension of 768 and 12 attention heads, whereas the huge+ variant consists of 32 transformer layers with a hidden dimension of 1280 and 20 attention heads, and uses SwiGLU as its feed-forward network.

\subsection{Pretraining objectives}
We adopt a discriminative self-supervised objective combining image-level and patch-level supervision. Specifically, the overall loss consists of a DINO loss~\cite{caron2021emerging} applied at the image level and an iBOT loss~\cite{zhou2021ibot} applied at the patch level, complemented by additional regularization terms.

At the image level, the DINO objective aligns the outputs of a student and a teacher network operating on different views of the same image. Let $p_s$ and $p_t$ denote the softmax-normalized outputs of the student and teacher  for a given view, respectively. $G$ and $L$ correspond to the sets of global and local views of an input image, respectively. The loss is defined as shown in Eq.~\ref{eq:dino}:
\begin{equation}
\label{eq:dino}
\mathcal{L}_{\mathrm{DINO}} = - \sum_{g \in G} \sum_{l \in L} p_t(g) \log p_s(l).
\end{equation}

At the patch level, the iBOT objective, shown in Eq.~\ref{eq:ibot}, extends this formulation to masked image modeling. A subset of patches is masked in the student input, while the teacher processes the full image. The student predictions for masked tokens are trained to match the corresponding teacher outputs, where $\mathcal{M}$ denotes the set of masked patches:
\begin{equation}
\label{eq:ibot}
\mathcal{L}_{\mathrm{iBOT}} = - \sum_{g \in G} \sum_{l \in L} \sum_{i \in \mathcal{M}} p_t^{(i)}(g) \log p_s^{(i)}(l).
\end{equation}

The final objective combines both terms, encouraging consistency at both global and local levels. Additional regularization such as feature decorrelation~\cite{sablayrolles2018spreading} is also applied denoted as $\mathcal{L}_{\mathrm{reg}}$, depicted in Eq.~\ref{eq:final_loss}:
\begin{equation}
\label{eq:final_loss}
\mathcal{L} = \mathcal{L}_{\mathrm{DINO}} + \mathcal{L}_{\mathrm{iBOT}} + 0.1 \cdot \mathcal{L}_{\mathrm{reg}}.
\end{equation}

\subsection{Post-Training refinement}
Following the established DINOv3 pretraining paradigm~\cite{simeoni2025dinov3}, we apply a post-training refinement stage consisting of resolution scaling and Gram anchoring. Prolonged self-supervised training can sometimes degrade patch-level features, reducing performance on dense prediction tasks. As demonstrated by DINOv3, this degradation is mitigated by a high-resolution adaptation step, in which the model is exposed to inputs at varying and higher resolutions. This enables the model to better capture fine-grained spatial details and improves robustness to resolution changes at inference time~\cite{simeoni2025dinov3}. 

In conjunction, we apply Gram anchoring, a regularization objective that enforces consistency between feature correlations of the student and a teacher model. Concretely, it aligns the Gram matrices of patch features, thereby preserving spatial structure and preventing the collapse of dense representations. This is particularly important at high resolutions, where maintaining coherent local features becomes more challenging~\cite{simeoni2025dinov3}.

Together, these refinements restore patch-level consistency and significantly enhance performance on dense tasks, while preserving strong global representations.

\subsection{Downstream task evaluation}

To comprehensively evaluate the learned representations, we consider five distinct surgical task categories across a total of 15 benchmarks. These tasks span complementary aspects of surgical understanding, including temporal workflow analysis, spatial perception, relational reasoning, and clinically grounded assessment. Importantly, they cover multiple procedures with diverse visual characteristics and workflow structures, enabling a systematic study of generalization across both tasks and domains.

Evaluating across multiple tasks and procedures is critical for two reasons. First, it allows us to assess whether learned representations capture transferable surgical knowledge beyond a single task or dataset. Second, it provides a controlled setting to analyze how label efficiency varies as a function of task structure, rather than being confounded by dataset-specific properties. To measure label efficiency in training-data subset experiments, we use three independent random video splits and report the mean $\pm$ standard deviation across splits.

\paragraph{Surgical Workflow Recognition}
We evaluate temporal understanding at two levels of granularity: coarse-grained surgical phases and fine-grained procedural steps. Phase recognition captures high-level stages of a procedure (e.g., preparation, dissection, closure), whereas step recognition focuses on more detailed sub-actions within each phase, requiring finer temporal discrimination.

For phase recognition, we consider Cholec80~\cite{twinanda2016endonet} (cholecystectomy), GraSP~\cite{ayobi2025pixel} (prostatectomy), MultiBypass140~\cite{lavanchy2024challenges}, comprising StrasBypass70 (gastric bypass, Strasbourg, France) and BernBypass70 (gastric bypass, Bern, Switzerland), AutoLaparo~\cite{wang2022autolaparo} (hysterectomy), and PmLR50~\cite{guo2025surgical} (liver resection). For step recognition, we evaluate on the respective steps annotations of StrasBypass70, BernBypass70, and GraSP.

This distinction between phases and steps enables us to study how label efficiency varies with temporal granularity, as step recognition typically requires more precise temporal localization and finer semantic understanding.

\paragraph{Surgical Semantic Segmentation}
To assess spatial perception, we evaluate semantic segmentation on Endoscapes-Seg50~\cite{murali2023endoscapes}, consisting of 50 videos with annotations for six semantic classes, and GraSP~\cite{ayobi2025pixel}, which provides pixel-wise annotations for seven surgical tools. These tasks require dense spatial predictions and provide strong supervision signals at the pixel level, enabling analysis of label efficiency under highly localized supervision.

\paragraph{Triplet Recognition}
We evaluate relational reasoning via triplet recognition, capturing interactions between surgical instruments, actions, and anatomical structures. For this we consider CholecT50~\cite{nwoye2022rendezvous}, which defines 100 unique instrument–verb–tissue combinations for cholecystectomy, and ProstaTD~\cite{chen2025prostatd} for prostatectomy with 89 interaction classes. These tasks require joint reasoning over multiple entities and represent a higher level of semantic abstraction compared to purely spatial or temporal tasks.

\paragraph{Critical View of Safety (CVS)}
We evaluate clinically grounded understanding using the Critical View of Safety (CVS) task on the Endoscapes dataset~\cite{murali2023endoscapes}. CVS assessment is based on three criteria defined by Strasberg~\cite{strasberg2010rationale}, which must be satisfied to ensure safe dissection during cholecystectomy. Each frame is annotated independently by three experts, and final labels are obtained via majority voting. This task requires high-level clinical reasoning and is particularly challenging due to its abstract and safety-critical nature.

\paragraph{Evaluation metrics}
We use task-appropriate evaluation metrics for each category. For phase and step recognition, we report the F1 score, which captures the balance between precision and recall in temporally structured predictions. For semantic segmentation, we use mean Intersection over Union (mIoU), a standard metric for evaluating pixel-wise overlap between predicted and ground-truth masks. For triplet recognition and CVS assessment, we report mean Average Precision (mAP), which reflects performance in multi-label and detection-style settings where multiple interactions or criteria may be present simultaneously and are threshold independent.

\paragraph{Normalized area under the curve (nAUC)}
To compactly quantify label efficiency across tasks and datasets, we use the normalized area under the curve (nAUC). The metric summarizes model performance as a function of annotation budget. For each task, performance is normalized by the maximum value achieved across all models. We then compute the area under the resulting performance curve using trapezoidal integration and normalize it by the range of the input domain, yielding a value in $[0,1]$. Higher nAUC indicates higher performance relative to the task-specific best across the annotation budget, reflecting greater overall label efficiency.

Let $\{(x_i, y_i)\}_{i=1}^n$ denote performance $y_i$ measured at training set sizes $x_i$, sorted such that $x_1 < \dots < x_n$. We normalize performance as shown in Eq.~\ref{eq:y_norm}:
\begin{equation}
    \label{eq:y_norm}
    \tilde{y}_i = \frac{y_i}{y_{\max}}, \quad \text{where } y_{\max} = \max_{1 \leq j \leq n} y_j.
\end{equation}

The normalized area under the curve (nAUC) is computed using trapezoidal integration displayed in Eq.~\ref{eq:nauc}:
\begin{equation}
\label{eq:nauc}
\mathrm{nAUC}
=
\frac{1}{x_n - x_1}
\sum_{i=1}^{n-1}
\frac{\tilde{y}_i + \tilde{y}_{i+1}}{2}
\left(x_{i+1} - x_i\right).
\end{equation}

To emphasize low-label regime learning behaviour, we additionally compute a low-label regime nAUC over the first two points $\tilde{y}_1$ and $\tilde{y}_2$ representing the performance at 1 and 3 training videos respectively, shown in Eq.~\ref{eq:low_nauc}. Higher low-label regime nAUC indicates stronger performance when only a small number of labeled videos are available:
\begin{equation}
\label{eq:low_nauc}
\mathrm{nAUC}_{\text{low-label}}
=
\frac{1}{x_2 - x_1}
\cdot
\frac{\tilde{y}_1 + \tilde{y}_2}{2}
\left(x_2 - x_1\right)
=
\frac{\tilde{y}_1 + \tilde{y}_2}{2}.
\end{equation}

\subsection{Implementation details}

\paragraph{Pretraining}

We perform large-scale self-supervised pretraining for both model variants, $\mathrm{ViT}{\text{-Base}}$ (SURGE-B) and $\mathrm{ViT}{\text{-Huge+}}$ (SURGE-H), on the SurgSpectrum-30M+. Pretraining is conducted for 180k iterations with a batch size of 2048, corresponding to approximately 12 epochs over the dataset.

Following this stage, we apply an additional post-refinement phase for 30k iterations with a reduced batch size of 64. This stage incorporates multi-resolution adaptation and Gram anchoring to further improve representation quality.

During multi-resolution training, we employ two global crops at resolutions of $512 \times 512$ and $768 \times 768$, and three local crops for the student network at $112 \times 112$, $168 \times 168$, and $336 \times 336$. Optimization is performed using a base learning rate of $4 \times 10^{-4}$ and a weight decay of 0.04.

\paragraph{Linear probing evaluation}

To evaluate the quality of the learned representations, we adopt a linear probing protocol across all downstream tasks. For image-level prediction tasks (i.e., all except semantic segmentation), we follow the protocol used in ~\cite{simeoni2025dinov3} by constructing a global representation via concatenating the $\mathrm{[CLS]}$ token with the average pooled patch tokens, which is then passed to a single-layer MLP for task-specific prediction.

For semantic segmentation, we directly use the patch-level features and apply a linear classification head to obtain dense predictions, which are subsequently upsampled via bilinear interpolation to the original image resolution. For convolutional architectures, we apply an equivalent strategy by operating on the final feature maps.

During linear probing, the pretrained backbone remains frozen and only the task-specific head is optimized. To ensure fair comparison, we perform a systematic hyperparameter search over learning rates for each task and dataset over the respective validation sets, and apply the same protocol consistently across all evaluated methods, including both our models and the considered baselines. Learning rates were selected from the following set based on validation performance:
[
$10^{-5}$, $2\times10^{-5}$, $5\times10^{-5}$, $10^{-4}$, $2\times10^{-4}$, $5\times10^{-4},10^{-3}$, $2\times10^{-3}$, $5\times10^{-3}$, $10^{-2}$, $2\times10^{-2}$, $5\times10^{-2}$, $10^{-1}$
]. 

This protocol isolates the quality of the learned representations and enables a controlled comparison of label efficiency across tasks without confounding effects from task-specific fine-tuning.

\paragraph{Higher parameter evaluation}

In addition, we consider a higher-capacity adaptation setting in which a subset of the pretrained backbone layers is fine-tuned together with the task-specific head. We perform this experiment only for SURGE-H, which achieved the best performance in the linear probing experiments, and conduct a hyperparameter search over the number of final layers to fine-tune ($k \in \{{1, 3, 6, 9, 12, 15}\}$) and the learning rate separately for each of the two datasets described in section~\ref{higher-param}. The remaining backbone layers are kept frozen, and the best configuration on the respective validation set is used for the final evaluation on the corresponding test set.

\section*{Data availability}
The pretraining corpus used in this study, \textbf{SurgSpectrum-30M+}, is a large-scale aggregation of both public and private surgical video sources. Publicly available components, including datasets such as Cholec80~\cite{twinanda2016endonet}, ProstaTD~\cite{chen2025prostatd}, SurgVu24~\cite{zia2025surgical}, OCHRA rectal cancer surgery~\cite{ghamrawi2025rectal}, and many more (see section ~\ref{sec:pretrain_data}) can be accessed through their respective official repositories. The private portions of the corpus consist of institutional data that cannot be publicly released due to patient privacy regulations and institutional data-sharing agreements. For downstream evaluation, all datasets (Cholec80, CholecT50, GraSP, Endoscapes, etc.) are publicly available through their original providers.

\section*{Code availability}
To support the reproducibility of our findings and facilitate further research in surgical foundation models, we provide the pretrained model weights for \textbf{SURGE-B and SURGE-H}, along with the necessary loading-, inference-,and evaluation scripts on all downstream tasks. The implementation is based on PyTorch and is available at \url{https://github.com/CAMMA-public/SURGE}.

\section*{Acknowledgements}

This work was partially supported by French state funds managed by the ANR under Grants ANR-22-FAI1-0001 (project DAIOR), ANR-10-IAHU-02 (IHU Strasbourg), ANR-23-IACL-0004 (AI Cluster Grand Est ENACT) and by the Interdisciplinary Thematic Institute HealthTech (ITI 2021-2028 program of the University of Strasbourg, CNRS and Inserm) via the IdEx Unistra (ANR-10-IDEX-0002) and SFRI (STRATUS project, ANR-20-SFRI-0012). This work has also received funding from the European Union (ERC, CompSURG, 101088553). Views and opinions expressed are however those of the authors only and do not necessarily reflect those of the European Union or the European Research Council. Neither the European Union nor the granting authority can be held responsible for them.

\section*{Author contributions}
F.S., N.N. and N.P. conceived and designed the work. F.S. contributed to the technical implementation and conducted experiments. L.A. and P.M. provided clinical expertise and collected the data for the self-supervised pretraining. The members of the CAMMA International Surgical Partners provided clinical expertise and access to proprietary datasets. F.S., L.A., V.S., P.M., and N.P. contributed to the drafting and revising of the manuscript. All authors have received and reviewed the manuscript. N.N. and N.P. supervised the research. 

\section*{Competing interests}
P.M. and N.P. are co-founders and shareholders of Scialytics. The remaining authors declare no competing interests.

\section*{CAMMA International Surgical Partners}

\textbf{Members}

Luigi Boni$^{1}$, Marta Goglia$^{2}$, Gianfranco Silecchia$^{2}$, Andrea Balla$^{3}$, Salvador Morales-Conde$^{3}$, Giovanni Guglielmo Laracca$^{4}$, Claudio Fiorillo$^{5}$, Giuseppe Quero$^{5}$, Vincenzo Tondolo$^{5}$, Ludovica Baldari$^{1}$, Elisa Cassinotti$^{1}$, Ludovica Guerriero$^{6}$, Isacco Montroni$^{7}$, Anna Fagotti$^{8}$, Alice Zampolini$^{8}$, Riccardo Oliva$^{8}$, Xin Wang$^{9}$, Matteo Pavone$^{8}$, Nicolò Bizzarri$^{8}$, Lise Lecointre$^{10}$
\\

\textbf{Affiliations}

1. Department of General \& Minimally Invasive Surgery
Fondazione IRCCS - Ca' Granda - Ospedale Maggiore Policlinico di Milano, Milan, Italy \\
2. Department of Medical-Surgical Sciences and Translational Medicine, Faculty of Medicine and Psychology, Sapienza University of Rome, Italy \\
3. University Hospital Virgen Macarena, University of Sevilla, Seville, Spain \\
4. Sant’Andrea Hospital, Sapienza University, Rome, Italy \\
5. Policlinico Universitario Agostino Gemelli IRCCS, Rome, Italy \\
6. Monaldi Hospital, AORN dei Colli, Naples, Italy \\
7. IRCCS Fondazione Istituto Nazionale dei Tumori, Division of Colon and Rectal Cancer Surgery, Milan, Italy \\
8. Gynecologic Oncology Unit, Fondazione Policlinico Universitario A. Gemelli IRCCS, Università Cattolica del Sacro Cuore, Rome, Italy \\
9. West China Hospital of Sichuan University, Chengdu, China \\
10. Department of Gynecologic Surgery, University Hospitals of Strasbourg, Strasbourg, France; ICube, Laboratory of Engineering, Computer Science and Imaging, Department of Robotics, Imaging, Teledetection and Healthcare Technologies, University of Strasbourg, CNRS, UMR 7357, Strasbourg, France \\

\bibliographystyle{unsrt}
\bibliography{arxiv}

\newpage
\setcounter{table}{0}
\renewcommand{\thetable}{\arabic{table}}
\captionsetup[table]{labelformat=simple, labelsep=period, name={eTable}}

\newcommand{\CI}[2]{[#1--#2]} 
\newcommand{\minu}{\,\mathrm{min}} 
\newcommand{\spermin}{\,\mathrm{s/min}} 

\clearpage

\end{document}